\documentclass[10pt,twocolumn]{article}
\usepackage{cvpr}
\usepackage{times}
\usepackage{epsfig}
\usepackage{wrapfig, blindtext}
\usepackage{graphicx}
\usepackage{amsmath}
\usepackage{cite}
\usepackage{amssymb}
\usepackage{fancyhdr}
\usepackage{parskip}
\usepackage{lettrine}
\usepackage{tikz}
\usetikzlibrary{arrows}
\usetikzlibrary{arrows.meta}
\usepackage{color}
\usepackage{xifthen}
\newcommand{\comment}[1]{}
\usepackage[breaklinks=true,bookmarks=false]{hyperref}

\cvprfinalcopy 
\def\cvprPaperID{****} 
\begin{document}
\title{Cooperation Is All You Need}


%

\newcommand*\samethanks[1][\value{footnote}]{\footnotemark[#1]}
\author{Ahsan Adeel\thanks{CMI Lab, University of Stirling, Stirling, UK.}  \thanks{deepCI.org, Parkside Terrace, Edinburgh, UK.  \\
Email: ahsan.adeel@deepci.org}
\and Junaid Muzaffar \samethanks[1]
\and Fahad Zia \samethanks[1]
\and Khubaib Ahmed \samethanks[1]
\and Mohsin Raza \samethanks[1]
\and Eamin Chaudary  \samethanks[1]
\and Talha Bin Riaz \samethanks[1]
\and Ahmed Saeed \samethanks[1]
}

\maketitle

\begin{abstract} 

Going beyond ‘dendritic democracy’, we introduce a ‘democracy of local processors’, termed Cooperator. Here we compare their capabilities when used in permutation-invariant neural networks for reinforcement learning (RL), with machine learning algorithms based on Transformers, such as ChatGPT. Transformers are based on the long-standing conception of integrate-and-fire ‘point’ neurons, whereas Cooperator is inspired by recent neurobiological breakthroughs suggesting that the cellular foundations of mental life depend on context-sensitive pyramidal neurons in the neocortex which have two functionally distinct points. We show that when used for RL, an algorithm based on Cooperator learns far quicker than that based on Transformer, even while having the same number of parameters.
\end{abstract}

\textbf{Introduction}
Transmitting information when it is relevant but not otherwise, is the fundamental capability of the biological neuron \cite{phillips2023cooperative}: but how does the neuron know what is relevant and what is not? The literature \cite{phillips2017cognitive} suggests that one of the functions of arousal and attention is to increase signal-to-noise ratio (SNR), however, knowing what is relevant (signal) and what is irrelevant (noise) is a difficult problem. For example, information relevant to one brain region could be irrelevant to other regions \cite{phillips2017cognitive}. \\
In the literature, scientists have proposed several bio-inspired attention mechanisms for artificial neural nets (ANNs) \cite{guo2022attention}, one of the most popular is Transformer \cite{vaswani2017attention}---the backbone of ChatGPT. However, existing attention mechanisms are based on the conception of integrate-and-fire `point' neurons \cite{hausser2001synaptic,burkitt2006review} that integrate all the incoming synaptic inputs in an identical way to compute a net level of cellular activation, also known as `dendritic democracy (DD)'. \\
Although DD allows deep nets to learn the representation of information with multiple levels of abstraction, it disregards the importance of cooperation between neurons i.e., individual neurons transmit information regardless of its relevance to the neighbouring neurons. This leads to the feed-forward (FF) transmission of conflicting messages, making learning difficult and increasing energy usage \cite{adeel2022context, adeel2022unlocking}.\\
\begin{figure} 
	\centering
	\includegraphics[trim=0cm 0cm 0cm 0cm, clip=true, width=0.5\textwidth]{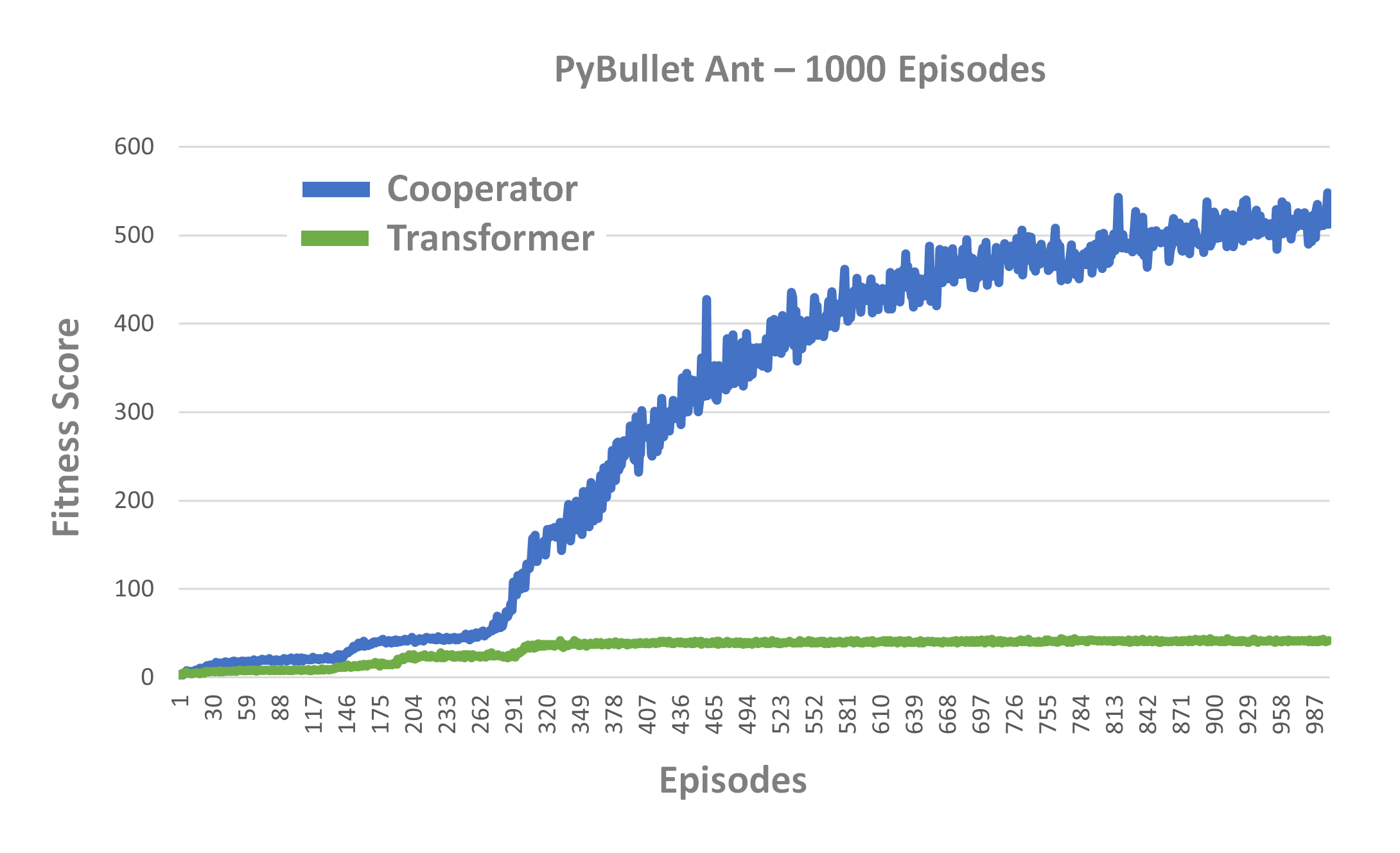}
	\caption{Permutation invariant RL agent (PyBullet Ant) adapting to sensory substitutions: Cooperator vs Transformer  \cite{vaswani2017attention, tang2021sensory}. In a fair comparison, with the same number of parameters, Cooperator learns far quicker than Transformer. See demo: will disclose after double-blind review.}
	\label{l5pc}
 \vspace{-1em}
\end{figure}
Recent neurobiological breakthroughs \cite{larkum1999new, larkum2013cellular, phillips2017cognitive} have revealed that two-point layer 5 pyramidal cells
(L5PCs) in the mammalian neocortex use their apical inputs as context to modulate the transmission of coherent feedforward (FF) inputs to their basal dendrites. These studies, including \cite{major2013active, phillips2023cooperative, ramaswamy2015anatomy,larkum2022dendrites, adeel2020conscious, kording2000learning, SchumanAnnual, poirazi2020, larkum2018perspective, kay1998contextually, kay2020contextual, kay2022comparison} have also devised context-sensitive neuro-modulatory transfer functions that motivate the transmission of information that is coherent. However making receptive field (RF) (or FF input) necessarily the driving force, has failed to produce promising results for complex real-world problems. Although a single single two-point neuron with apical dendrites can solve the exclusive-or (XOR) problem that is solvable only by multiple layers of conventional artificial point neurons \cite{poirazi2020}, how they perform their magic at scale has, until now, remained enigmatic.\\
Going beyond DD, we address this long-standing issue by introducing ‘democracy of local processors (DoLP)’, termed Cooperator. Rather than FF information being the driving force behind neural output, DoLP enables local processors to overrule the dominance of RF and awards more authority to the contextual information coming from the neighbouring neurons \cite{adeel2022context, adeel2022unlocking}. This context-sensitivity in two-point neurons amplifies or suppresses the transmission of FF information when the context shows it to be relevant or irrelevant respectively. See our spiking context-sensitive two point neurons simulation with burst-dependent synaptic plasticity \cite{payeur2021burst}: will disclose after double-blind review. \\
At a granular level, the context-sensitive processor uses context to estimate whether its perception about the RF aligns with the majority of neighboring processors; if it does, the transmission of RF is amplified else suppressed. This context-sensitive neural information processing is cooperative in that it seeks to maximize agreement between the active neurons, thus reducing the transmission of conflicting information. \\
DoLP may contains aspects of the highly influential `biased competition’ as a theory of attention and normalization \cite{reynolds2009normalization} and of the recurrent amplification \cite{ carandini2012normalization} for which the biophysical and cellular bases are outlined in \cite{phillips2023cooperative}. \\In \cite{adeel2022context, adeel2022unlocking}, researchers showed that such context-sensitive neural information processing can process large-scale complex real-world data far more effectively and efficiently than state-of-the-art point neurons-inspired deep nets. Here we show that this approach is capable of learning extremely fast compared to Transformer when used in permutation-invariant neural networks for RL (Figure 1) \cite{tang2021sensory}. 

\begin{figure*} 
	\centering
	\includegraphics[trim=0cm 0cm 0cm 0cm, clip=true, width=1\textwidth]{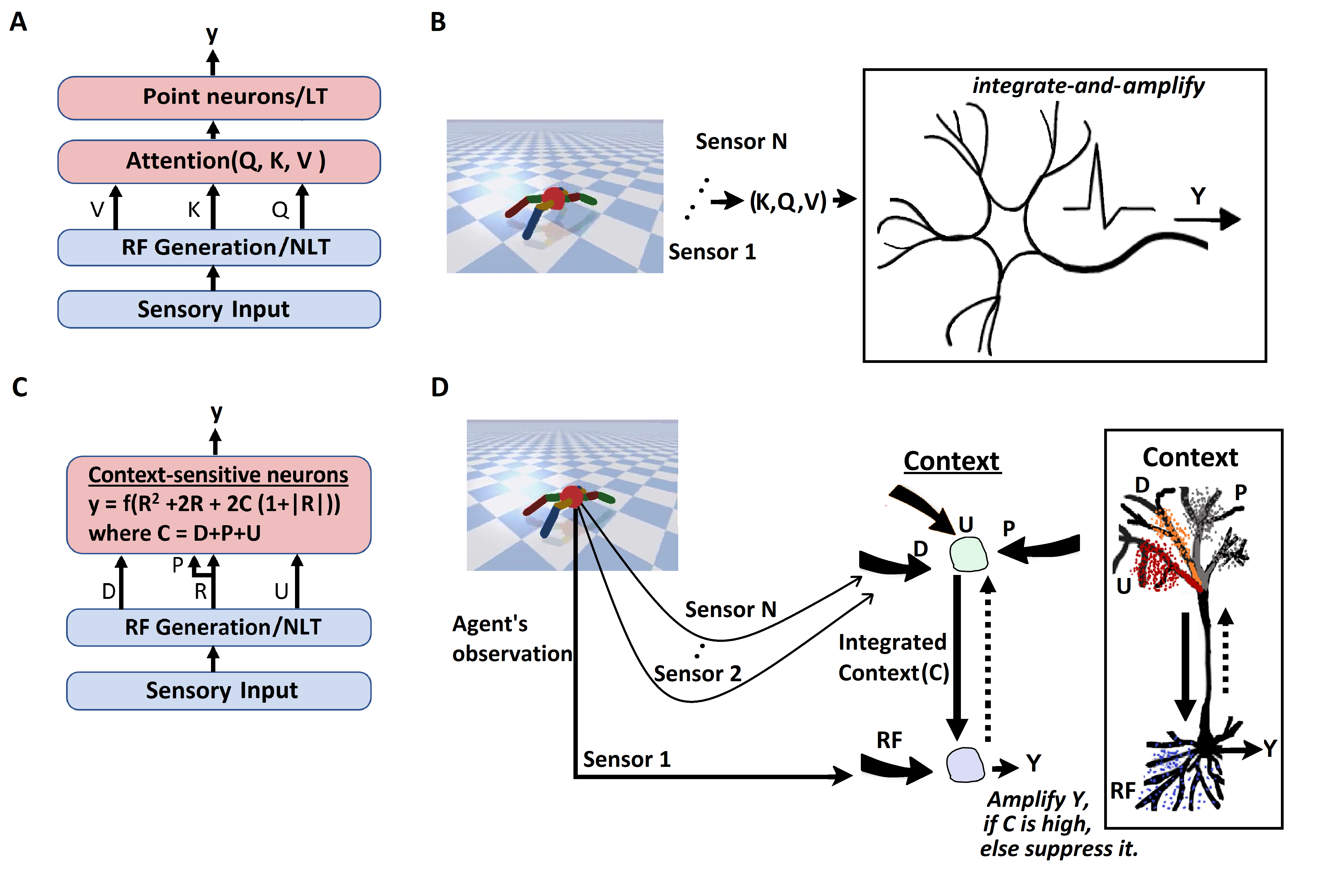}
	\caption{\textbf{(A)}: Point neuron-based Transformer. Scaled Dot-Product Attention or Multi-Head Attention \cite{vaswani2017attention} used to model permutation invariant RL agent \cite{tang2021sensory}. \textbf{(B)} A simple representation of Point neuron-based Transformer for permutation invariant PyBullet Ant RL agent. The point neurons simply sum up all the inputs with an assumption that they have the same chance of affecting the neuron's output. \textbf{(C)} Context-sensitive neuron-based Cooperator used to model permutation invariant RL. \textbf{(D)} Functional depiction of a context-sensitive neuron with two points of integration whose contextual integration zone receives proximal context (P) from neighboring sensory 1 neurons, distal context (D) from more distant parts of the network (sensory neurons 2-N), and universal context (U) representing Q. The integrated context (C) is used as an average opinion of the neighboring neurons to decide whether to transmit the information or not. Higher the value of C, higher the probability of transmitting the information. For more details, see \cite{adeel2022context,adeel2020conscious}.}
	\label{l5pc}
 \vspace{-1em}
\end{figure*}

\textbf{Transformer vs. Cooperator}\\
Figure 2(A) shows state-of-the-art Transformer's Scaled Dot-Product Attention or Multi-Head Attention that uses three different representations of RF via linear transformations (LTs) or non-linear transformations (NLTs), representing Query (Q), Key (K) and Value (V) matrices, given as \cite{vaswani2017attention}:
\begin{equation}
Attention(Q, K, V) = f(QK^TV)   
\end{equation}

An equivalent point neuron representation of Transformer for permutation invariant RL agent (PyBullet Ant) adapting to sensory substitutions is shown in Figure 2(B) \cite{tang2021sensory}. The point neurons integrate all the incoming sensory streams in an identical way i.e., simply summing up all the inputs with an assumption that they have the same chance of affecting the neuron's output \cite{hausser2001synaptic}. \\
In contrast, the proposed Cooperator network (Figure 2(C)) uses a cooperative context-sensitive neural information processing mechanism \cite{adeel2022context} in which cooperative context-sensitive neural processors (Figure 2(D)) receive two functionally distinct sets of inputs. One set provides the input about which the neuron transmits information: RF. The other set provides opinion of the neighboring neurons about the RF as context. These processors use  context to amplify or attenuate the transmission of relevant or irrelevant information, respectively. Specifically, here the neuron that is sensitive to Sensor 1, receives information from the neighbouring neurons of the same neural net (NN) as proximal context (P), from distal neurons sensitive to Sensors 2-N (where N represents represents total number of Sensors) in more distant parts of the network as distal context (D), and all possible pairs of input as universal context (U). The neuron uses integrated context (C) via asynchronous modulatory transfer function eq (2) \cite{adeel2022context} to selectively amplify and suppress the FF transmission of the relevant and irrelevant Sensor 1 information, respectively. Same applies to all other neurons. 

This new asynchronous modulatory transfer function (AMTF), termed `Cooperation Equation' can be defined as: 
\begin{equation}
Cooperation(R, C) = f(R^2 + 2R + 2C(1+|R|))  
\end{equation}
This cooperation equation enforces ‘democracy of local processors' that can over-rule outliers. In this equation, C is the driving force that decides whether to amplify or suppress the transmission of information \cite{adeel2022context}. Specifically, individual neurons use C as a `modulatory force' to push the neuron's output to the positive side of the activation function (e.g., rectified linear unit (ReLU)) if R is relevant, otherwise to the negative side. In essence, C can discourage or encourage amplification of neural activity if R is strong or weak, respectively \cite{adeel2022context}. This mechanism enhances cooperation and seeks to maximise agreement between the active neurons. 

Below are the alternative well-established AMTFs proposed by others \cite{kay1998contextually, phillips2023cooperative}. In these AMTFs ($T_{Ms}$) eq (3-6), R is the driving force i.e., if R is absent or strong, C has no role to play. 

\begin{equation}
T_{M1}(R, C) = \frac{1}{2}R(1+exp(RC))  
\end{equation}

\begin{equation}
T_{M2}(R, C) = R+RC  
\end{equation}

\begin{equation}
T_{M3}(R, C) = R(1+tanh(RC))  
\end{equation}

\begin{equation}
T_{M4}(R, C) = R(2^{RC})  
\end{equation}

In the multisensory RL case used here, R, P, and D are functions of the Sensors 1-N i.e., input $x$ $\epsilon$ $R^N$ (e.g., any LT or NLT) and U is the output of positional encoding \cite{vaswani2017attention} matched to the dimensions of R, P, and D. For permutation invariance (PI), U is independent of input $x$ such that permuting $x$ only effects P and D but not U, which enables the output to be PI \cite{tang2021sensory}. As explained comprehensively in \cite{tang2021sensory}, the individual sensory inputs 1-N or observations $O_t^i$, \textit{i=1, 2, ... N} along with the previous action $a_{t-1}$ passes through a NN module in an arbitrary order such that each NN has partial access to agent's obervation at time \textit{t} and $i^{th}$ neuron can only see the $i^{th}$ component of the observation $O_t[i]$, computing $f_R(O_t[i]$, $a_{t-1})$ and $f_D(O_t[i])$. The overall operation can be described using eq(7-10):

\begin{equation}
R(O_t,a_{t-1})
 =
  \begin{bmatrix}
   f_R(O_t[1], a_{t-1}) \\
   ...\\
 f_R(O_t[N], a_{t-1}) 
   \end{bmatrix} \in  \mathbb{R}^{N \times d_{f_R}} 
\end{equation}
\begin{equation}
D(O_t)
 =
  \begin{bmatrix}
   f_D(O_t[1]) \\
   ...\\
 f_D(O_t[N]) 
   \end{bmatrix} \in  \mathbb{R}^{N \times d_{f_D}} 
\end{equation}

\begin{multline}
    m_t = ReLU(R(O_t, a_{t-1})^2 + \\ 2R(O_t, a_{t-1}) + 2C(1+|R(O_t, a_{t-1}))|)
\end{multline}

\begin{equation}
    C = P+D+U
\end{equation}

Where P =  $f_P(R)$ and U is the function of positional encoding.

\begin{figure*} 
	\centering
	\includegraphics[trim=0cm 0cm 0cm 0cm, clip=true, width=1\textwidth]{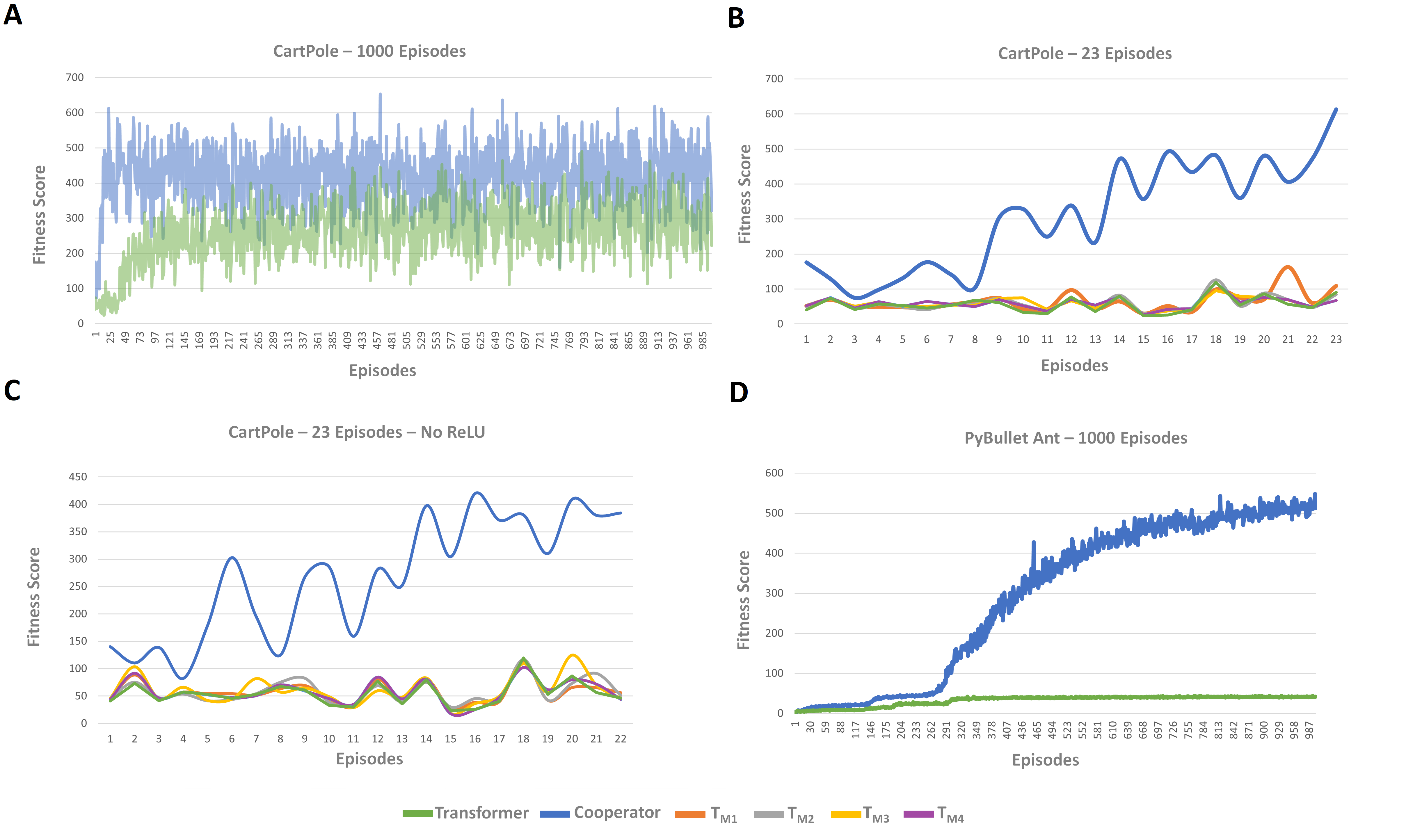}
	\caption{Training Results: In both Cart-Pole and PyBullet problems, Cooperator with the same architecture and number of parameters, learns far quicker than Transformer and previously proposed neuro-modulatory functions. In \cite{tang2021sensory}, the authors only presented testing results, here we present both training and testing results. See demo: will disclose after double-blind review. }
	\label{l5pc}
 \vspace{-1em}
\end{figure*}


\textbf{Results}

Due to limited processing power available, we could conveniently experiment with two different RL environments, Cart-pole swing up and PyBullet Ant for 10K and 1K iterations, respectively. The architectures of the policy networks, training methods, AttentionNeuron layers, and hyperparameters in all agents are same as used in \cite{tang2021sensory}. Results presented here are generated using the code provided in \cite{tang2021sensory}, which is also the baseline. \\
Figure 3 depicts results for Cart-pole and PyBullet Ant scenarios. It was observed that the context-sensitive neuron-driven agent learned the tasks far more quickly than the state-of-the-art Transformer based PI agents (baseline). Furthermore, the previously proposed context-sensitive neuro-modulation transfer functions (3-6) performed comparably to the baseline Transformer model. Specifically, in the Cart-pole problem, Cooperator in less than 23 episodes converges to the highest fitness score, crossing 600. In contrast, the baseline learns far slower, and reaches to the fitness score of 100 in 23 episodes, and remains below 500 mark in 1K episodes. In PyBullet Ant problem, the Cooperator learns even faster and crosses 500 mark in 1000 episodes. In contrast, the baseline and other TFs, never cross the fitness score of 100. Testing results for CartPole in Table 1 shows that cooperator trained over 1k-10K episodes achieve significantly higher fitness score with far less standard deviation, both in shuffled and unshuffled scenarios. Although for shuffled inputs Cooperator performed comparably to the baseline in 1K episodes, quickly jumped to the higher fitness score with less standard deviation in 5K episodes. However, in PyBullet Ant case, Cooperator outperformed in both shuffled and unshuffles scenarios. We are now training these models for 20k episodes and will report comparative results elsewhere in the future.

\begin{table} [htb!]
\centering
\caption{Cart-pole Test (trained over 1K, 5K, and 10K iterations). For each experiment, we report the average score and the standard deviation
from 1K test episodes.}
\begin{tabular}[t]{lcccc}
\hline
& & Iterations \\
\hline
\hline
& 1K  & 5K  & 10K \\
\hline

Transformer & 279$\pm$272 & 340$\pm$308 & 340$\pm$310 \\
Transformer (Shuffled) & 279$\pm$274 & 339$\pm$308 & 340$\pm$309 \\
Cooperator & 428$\pm$293 & 524$\pm$408 & 538$\pm$419 \\
Cooperator (Shuffled) & 267$\pm$248 & 508$\pm$408 & 536$\pm$417 \\
\hline
\end{tabular}
\end{table}%

\begin{table}[htb!]
\centering
\caption{PyBullet Ant test (trained over 1K episodes). For each experiment, we report the average score and the standard deviation
from 1K test episodes.}
\begin{tabular}[t]{lcc}
\hline
& ES & ES (shuffled)\\
\hline
Transformer & 121$\pm$53 & 30$\pm$241\\
Cooperator & 1170$\pm$35 & 280$\pm$124\\
\hline
\end{tabular}
\end{table}%
\textbf{Discussion}
\\Similar to the results presented in \cite{adeel2022context} for audio-visual speech processing, the results for RL presented here support our hypothesis that the fundamental weakness of state-of-the-art deep learning is its dependence on point neurons that inherently maximise the transmission of information irrespective of its relevance in the current context. In contrast, in the proposed cooperative context-sensitive neural information processing mechanism, neurons cooperate moment-by-moment with neighbouring neurons to amplify and suppress the transmission of relevant and irrelevant feedforward information, respectively. This mechanism ensures that the democracy of local processors prevails. \\
Although the convincing evidence presented in \cite{adeel2022context, adeel2022unlocking} showed that how context-sensitive neurons quickly evolve to become highly sensitive to a specific type of high-level information and `turn on' only when the received signals are relevant in the current context, leading to faster mutual information estimation, reduced neural activity, reduced energy consumption, and enhanced resilience,  the results presented here further endorse our radical point of view. 
In this study, Cooperator model consists of only one layer of two-point neurons followed by a simple policy network. Furthermore, the architecture, including the number of parameters, is the same as in \cite{tang2021sensory}. For CartPole and PyAnt simulations, please see: will disclose after double-blind review. We are currently training deeper models consisting of multiple layers of two-point neurons for Language Models. However, results for a deeper network applied to audio-visual speech processing are shown in \cite{adeel2022context, adeel2022unlocking}. For a 50-layered deep net, please see: will disclose after double-blind review.\\
The evidence on sensory substitution was one of many grounds for supposing that context-sensitive processing is central to cortical computation, as argued in \cite{phillips1997search}, and more recently supported in \cite{harris2015neocortical}. 
These results strongly support the cooperative context-sensitive views of neocortical function. It is worth mentioning that our algorithms are not neural models, but a demonstration that the cooperative context-sensitive  style of computing has exceptional big data information processing capabilities that could be implemented either in silicon, or in neural tissues.

\textbf{Acknowledgments}
This research was supported by the UK Engineering and Physical Sciences Research Council (EPSRC) Grant Ref. EP/T021063/1. We would like to acknowledge Professor Bill Phillips and Professor Leslie Smith from the University of Stirling, 
Professor Peter Konig from the University of Osnabruck, Professor Newton Howard from Oxford Computational Neuroscience, Dr Michael Spratling from King's College London, and Professor Michael Hausser from University College London for their help and support in several different ways, including reviewing our work, appreciation, and encouragement. \\

\textbf{Contributions}
AA conceived and developed the original idea, wrote the manuscript, and analyzed the results. AA, JM, KA, MR, FZ, EC, TBR, and AS performed simulations. \\

\textbf{Competing interests}
AA has a provisional patent application for the algorithm used in this paper. The other authors declare no competing interests.

\textbf{Data availability} The data that support the findings of this study are available on request.
\bibliographystyle{IEEEtran}
\bibliography{NATURE.bib}
\onecolumn 
\pagebreak 

\end{document}